\documentclass[12pt]{article}

\usepackage{booktabs}
\usepackage{siunitx}
\usepackage{amsmath}
\usepackage[ruled,vlined,linesnumbered]{algorithm2e}

\usepackage{etoolbox}
\usepackage{bm}

\usepackage{arydshln}

\makeatletter
\def\mathcolor#1#{\@mathcolor{#1}}
\def\@mathcolor#1#2#3{%
	\protect\leavevmode
	\begingroup
	\color#1{#2}#3%
	\endgroup
}
\makeatother

\usepackage{url}
\usepackage{setspace}

\usepackage[table,xcdraw]{xcolor}

\usepackage{graphicx}
\usepackage{multirow}

\graphicspath{ {./figures/} }

\usepackage{caption}
\usepackage{subcaption}

\usepackage{amsmath} 
\usepackage{amsfonts}
\usepackage{nicefrac}

\usepackage[a4paper, total={6in, 8in}]{geometry}

\usepackage[symbol]{footmisc}
\usepackage[
    colorlinks=true,
    linkcolor=blue,
    filecolor=blue,
    urlcolor=blue,
    citecolor=blue,
]{hyperref}

\makeatletter

\usepackage{xspace}   

\newcommand{\fm}{FlexiMIND\xspace}
\newcommand{\mind}{MIND\xspace}
\newcommand{\mindssc}{MIND-SSC\xspace}
\newcommand{\ca}{ConvexAdam\xspace}

\newcommand{\dinoreg}{DINO-Reg\xspace}
\newcommand{\flexict}{FlexiCT-2D\xspace}

\newcommand{\dirlab}{DIR-Lab 4DCT\xspace}
\newcommand{\acdc}{ACDC\xspace}
\newcommand{\abd}{Abdomen MR-CT\xspace}
\newcommand{\ctct}{CT-CT\xspace}
\newcommand{\mrmr}{MR-MR\xspace}
\newcommand{\mrct}{MR-CT\xspace}

\newcommand{\tre}{TRE\xspace}
\newcommand{\dice}{Dice\xspace}
\newcommand{\dsc}{DSC\xspace}
\newcommand{\sdlogj}{\mbox{SDlogJ}\xspace}

\newcommand{\mm}[1]{\SI{#1}{\milli\meter}}

\newcommand{\pct}[1]{\SI{#1}{\percent}}

\usepackage[
    backend=biber,
    style=numeric,
    sorting=none,
    maxbibnames=6,
    minbibnames=6
]{biblatex}

\begin{document}
	\begin{center}	
		\textbf{\large {FlexiMIND: Combined Pretrained Vision Features and Local Structure for Training-Free Deformable Medical Image Registration}} \\
		Mojtaba Safari$ ^1 $, Mingzhe Hu$ ^1 $, Shansong Wang$ ^1 $, Suyang Xi$ ^2 $, Joshua Lorenz$ ^1 $, Sean Pitroda$ ^1 $, Aytekin Oto$ ^{3, \ddagger} $, Ralph R. Weichselbaum$ ^{1, \ddagger} $, and Xiaofeng Yang$ ^{1, \ddagger} $ \\
    \end{center}
\begin{flushleft}
$ ^1 $Department of Radiation and Cellular Oncology, The University of Chicago, Chicago, IL, United States.\\	
$ ^2 $Department of Computer Science and Informatics, Emory University, Atlanta, GA, United States.\\
$ ^3 $Department of Radiology, The University of Chicago, Chicago, IL, United States.\\
\vspace{1 cm}
$ ^\ddagger  $\textbf{Corresponding Authors}: \\ \url{aoto@bsd.uchicago.edu} (AO);  \url{rweichselbaum@uchicagomedicine.org} (RW); \url{xfyang@uchicago.edu}  (XY)
\end{flushleft}

\newpage

\begin{abstract}
	\noindent\textbf{Objective.}  
	Deformable image registration remains challenging because corresponding anatomy may exhibit different intensity relationships. Hand-crafted modality-robust descriptors such as Modality-Independent Neighborhood Descriptor with Self-Similarity Context (\mindssc) and features from pretrained vision encoders provide distinct representations for correspondence estimation, but how they should be combined remains unclear. We investigate whether tighter integration improves registration over a sequential, scale-separated design.
	\noindent\textbf{Approach.}
	We introduce \fm, a training-free and label-free two-stage framework in which frozen \flexict encoder features establish contextual correspondence and \mindssc refines the deformation within \ca. Five alternative feature-coupling strategies were evaluated under a protocol across CT-CT (\dirlab 4DCT, 10 cases), MR-MR (\acdc, 50 patients), and MR-CT (Learn2Reg \abd, 8 pairs) registration. Performance was assessed using target registration error (\tre), Dice similarity coefficient (\dice), \sdlogj, and local folding.
	\noindent\textbf{Main results.}
    \fm achieved a \tre of \SI{1.27}{\milli\meter} on \dirlab, a DSC of $0.807$ on \acdc, and an organ-averaged DSC of $0.825$ on \abd. Across 15 coupling-strategy dataset comparisons, only four outperformed the reference method, with a maximum absolute improvement of $0.0043$ in Dice, whereas degradation reached \SI{20.6}{\percent} in \tre and \SI{37.9}{\percent} in Dice. Changes in registration accuracy were closely associated with changes in deformation-field regularity, and strategies that better preserved regularity generally remained closest to the reference performance. Landmark-level analysis showed that the pretrained encoder stage primarily reduced large correspondence errors rather than uniformly improving local precision.
	\noindent\textbf{Significance.}
    These findings support a scale-separated strategy for training-free multimodal registration, in which pretrained vision features establish large-scale correspondence and local self-similarity descriptors refine the deformation. The results also highlight deformation-field regularity when evaluating feature-coupling strategies. \fm requires neither task-specific training nor anatomical labels and generalizes across different modality pairings without retraining. Code is available upon acceptance at \url{https://github.com/mosaf/FlexiMIND}.
\end{abstract}

\newpage

\section{Introduction}
\label{sec:intro}

Medical image registration establishes spatial correspondence between images and is fundamental to applications such as longitudinal target assessment, radiation treatment planning and guidance, anatomical delineation, atlas-based analysis, and multimodal image fusion~\cite{chen2025regsurvey}. Deformable image registration extends this task to spatially varying anatomical differences by estimating a dense transformation between the fixed and moving images. Its accuracy therefore depends on establishing reliable anatomical correspondences throughout the image. This becomes particularly challenging when image appearance differs between the two images, especially in multimodal registration, where corresponding anatomy can exhibit substantially different intensity and contrast patterns. Classical similarity measures such as mutual information (MI) and normalized gradient fields (NGF) reduce the dependence on direct intensity correspondence~\cite{maes1997mi,haber2006ngf}. Another approach is to represent images in a feature space that is less sensitive to modality-dependent appearance while retaining sufficient anatomical information for correspondence estimation.

Hand-crafted modality-invariant descriptors have provided an established representation for this purpose. The modality independent neighborhood descriptor (\mind) characterizes local self-similarity rather than absolute image intensity~\cite{heinrich2012mind}, and its self-similarity context formulation, \mindssc, extends this representation by encoding relationships among neighboring patches~\cite{heinrich2013ssc}. Combined with optimization-based registration methods such as \ca~\cite{convexadam2021}, \mindssc has achieved competitive performance on public registration benchmarks without registration-specific training~\cite{hering2022learn2reg}. However, its local construction restricts the spatial context available for correspondence estimation. Structures with similar local appearance may therefore remain difficult to distinguish when correct correspondence requires broader anatomical context.

Features extracted from pretrained vision encoders offer a different representation of anatomical correspondence. Self-supervised vision transformers such as DINOv2 have learned contextual representations that transfer across imaging domains without task-specific fine-tuning~\cite{oquab2023dinov2}. DINO-Reg demonstrated that PCA-reduced DINOv2 features could be incorporated into a \ca-based optimization framework to establish correspondence without training a registration network on the target dataset~\cite{dinoreg2025}. In our previous work, we further investigated this direction using DINOv3 representations with test-time optimization of the deformation field~\cite{10.1117/12.3087870}. Although learning-based registration methods have also achieved strong performance~\cite{chen2022transmorph,fu2020deep}, their dependence on registration-specific training makes them methodologically different from the training-free setting considered in this study.

These two feature families encode anatomical information at different spatial scales. \mindssc preserves local three-dimensional structural information at relatively high spatial resolution, whereas pretrained encoder features can provide broader contextual information but are generally spatially coarser and, when extracted from two-dimensional slices, contain limited explicit through-plane information. These differences raise the possibility that the two representations could provide complementary correspondence information. However, previous studies have primarily established effectiveness these approaches independently, and it remains unclear whether combining the information provided by these distinct feature families actually improves deformable registration. Moreover, comparisons involving different representations often change other components of the registration pipeline, such as the solver, feature dimensionality, working resolution, preprocessing, or regularization. Improvements observed under such conditions cannot be attributed uniquely to feature coupling. Changes in the correspondence representation may also alter deformation-field regularity, which can independently influence landmark- and overlap-based measures of registration accuracy.

In this study, we investigated whether coupling \mindssc with pretrained encoder features improved deformable registration when the remaining components of the registration pipeline were controlled. Specifically, we addressed two questions: (1) whether coupling the two representations improved registration accuracy when the solver and experimental protocol were held fixed, and (2) whether changes in registration accuracy reflected additional correspondence information or were primarily associated with changes in deformation-field regularity. We developed a training-free framework that incorporated both representations within a common \ca-based optimization pipeline and evaluated alternative coupling strategies across \ctct, \mrmr, and \mrct registration. Here, \emph{training-free} indicates that no parameters were optimized using registration data, although pretrained encoder weights were frozen. \emph{Label-free} indicates that no task-specific anatomical labels were used during registration optimization.

The main contributions of this study are:
\begin{itemize}
	\item We develop a training-free deformable registration framework that combines pretrained vision-encoder features with \mindssc within a common optimization pipeline, without registration-specific training or task-specific anatomical labels.
	\item We systematically evaluate alternative feature-coupling strategies across \ctct, \mrmr, and \mrct registration while controlling the solver, preprocessing, working resolution, and evaluation protocol.
	\item We show that tighter coupling of \mindssc and pretrained encoder features does not consistently improve registration accuracy over the sequential cascade, indicating that their different representational properties do not necessarily translate into complementary correspondence information.
	\item We demonstrate that changes in registration accuracy are closely associated with deformation-field regularity, identifying regularity as an important factor when interpreting improvements attributed to feature coupling.	
\end{itemize}

\section{Materials and Methods}
\label{sec:method}

\subsection{Datasets and evaluation protocol}
\label{sec:data}

Three public datasets were used to evaluate \ctct, \mrmr, and \mrct registration. Within each dataset, all configurations used identical preprocessing, solver settings, working resolution, affine initialization, and evaluation procedures.

The \dirlab dataset~\cite{castillo2009dirlab} contains ten thoracic 4-D CT cases with 300 expert-defined landmark correspondences per case between the extreme respiratory phases. Landmark indices were converted to physical coordinates using the spatial metadata of their respective image phases. Registration accuracy was quantified using target registration error (\tre), defined as the mean Euclidean distance between corresponding fixed and transformed moving landmarks. Following the DIR-Lab convention, transformed landmarks were rounded to the nearest voxel center of the moving image before computing \tre. 

The \acdc dataset~\cite{bernard2018acdc} was evaluated using the 50 official testing subjects. End-systolic and end-diastolic cine MR volumes were used as registration pairs, with expert segmentations of the right-ventricular cavity, myocardium, and left-ventricular cavity. Registration accuracy was quantified using \dice after transforming the moving segmentation directly onto the native fixed label grid. Results are reported using subject-averaged \dice, defined as the mean across subjects. Organ-averaged \dice, defined as the mean of the per-organ averages, was computed separately where required for comparison with published results. Native voxel dimensions are approximately $1.64 \times 1.64 \times 10$\,\si{\milli\meter}, and registration was performed at \mm{1.0} isotropic resolution with a fixed registration direction across configurations.

The Learn2Reg \abd dataset~\cite{hering2023learn2reg} contains eight paired MR and CT volumes with annotations of the liver, spleen, right kidney, and left kidney. MR was treated as the fixed image and CT as the moving image. Registration was performed directly on the released $192 \times 160 \times 192$ grid at \mm{2} isotropic resolution without additional affine pre-alignment or post-registration resampling. Evaluation used the official Learn2Reg software and distributed task configuration at a fixed software revision. Subject-averaged and organ-averaged \dice were computed separately.

In addition to registration accuracy, deformation-field regularity was evaluated using the standard deviation of the log Jacobian determinant (\sdlogj). Let $J_u(x)=\mathrm{Id}+\nabla u(x)$ denote the Jacobian of the estimated transformation. Following the official Learn2Reg implementation~\cite{hering2023learn2reg},

\begin{equation}
	\mathrm{SDlogJ} = \left[\frac{1}{\mid \Omega_{\mathrm{eval}} \mid}\sum_{x\in\Omega_{\mathrm{eval}}}\left(g_u(x)-\overline{g_u}\right)^2	\right]^{1/2},	\qquad
	g_u(x)=	\log\left(\det J_u(x)+3\right),
    \label{eq:sdlogj}
\end{equation}
where $\Omega_{\mathrm{eval}}$ denotes the evaluation region and $\overline{g_u}$ is the mean of $g_u$ within that region. Higher \sdlogj indicates greater spatial variation in the deformation field. Local folding was additionally quantified as the proportion of voxels satisfying $\det J_u(x)\leq0$ within the same evaluation region.

Methods that were not reimplemented were included only as aggregate benchmark comparisons using values reported in their source publications and were excluded from paired analyses. The released \dinoreg implementation~\cite{dinoreg2025} was also evaluated under the study protocol using the same working grids and cached affine initialization as the other measured configurations. Its published \mrct result on \abd was additionally included as an external reference; no corresponding published results were available for \dirlab or \acdc.

Runtime was measured on one NVIDIA RTX 6000 Ada Generation GPU with 48\,GB memory.

All within-dataset comparisons were paired because each configuration was evaluated on the same image pairs. Uncertainty in the mean registration metrics was quantified using \pct{95} bias-corrected and accelerated bootstrap confidence intervals based on 4000 resamples. Analyses were performed using per-case measurements for \dirlab ($n=10$), \acdc ($n=50$), and \abd ($n=8$). The feature-coupling strategies were evaluated relative to the reference \fm configuration using the same case-wise accuracy and deformation-field metrics.

\subsection{Registration framework}
\label{sec:framework}

Let $I_{\mathrm{fix}}:\Omega\rightarrow\mathbb{R}$ and $I_{\mathrm{mov}}:\Omega'\rightarrow\mathbb{R}$ denote the fixed and moving images. Deformable registration seeks a transformation $\varphi:\Omega\rightarrow\Omega'$, represented by the displacement field $u(x)=\varphi(x)-x$, by minimizing

\begin{equation}
	\mathcal{E}(u) 	= 	\underbrace{ \int_{\Omega} \mathcal{D} \left[ \Phi_{\mathrm{fix}}(x), \Phi_{\mathrm{mov}}\left(x+u(x)\right) \right] \mathrm{d}x }_{\text{data term}} + \lambda \underbrace{ \int_{\Omega} \parallel\nabla u(x)\parallel^2 	\mathrm{d}x }_{\text{regularity term}}, 
    \label{eq:objective}
\end{equation}
where $\Phi$ denotes the representation used for correspondence estimation, $\mathcal{D}$ is the feature dissimilarity, and $\lambda$ controls deformation regularity. We considered two representations: the hand-crafted \mindssc descriptor, denoted $M^{12}$, and features extracted from a frozen pretrained vision encoder, denoted $F^q$.The complete registration framework is illustrated in \figurename~\ref{fig:pipeline}.

For the hand-crafted representation, shown in \figurename~\ref{fig:pipeline}(b), we used the self-similarity context formulation of \mind, \mindssc~\cite{heinrich2012mind,heinrich2013ssc}. For a voxel $x$, its twelve descriptor channels are defined as

\begin{equation}
	\mathbf{m}_k(x) = \exp\left(-\frac{D_k(x)}{V(x)}\right), \qquad k=1,\ldots,12, 
    \label{eq:mind}
\end{equation}
where $D_k(x)$ is the sum of squared differences between the patches associated with the $k$th neighborhood pair and $V(x)$ is the mean of the corresponding patch-difference values at $x$. Each descriptor was normalized by its maximum value, and the channels were stacked as

\begin{equation}
	M^{12} 	= 	[\mathbf{m}_1,\ldots,\mathbf{m}_{12}].
\end{equation}

The fixed and moving descriptors are denoted $M^{12}_{\mathrm{fix}}$ and $M^{12}_{\mathrm{mov}}$, respectively. \mindssc was computed at the full working resolution using patch radius $r=1$ and neighborhood dilation $d=2$ voxels, as illustrated in \figurename~\ref{fig:pipeline}(b).

For the pretrained representation, as shown in \figurename~\ref{fig:pipeline}(a), features were extracted independently from each axial slice using the frozen \flexict model~\cite{flexict2026}, a CT-pretrained vision encoder with patch size 8. Features from the last four transformer layers were concatenated to obtain $C_{\mathrm{in}}=3456$ channels. Each slice was resized to a fixed token grid before encoding, $80\times70$ tokens for \dirlab and \acdc and $112\times96$ for \abd. The slice-wise features were subsequently stacked and upsampled to the working registration grid to obtain $F_{\mathrm{fix}}$ and $F_{\mathrm{mov}}$. The encoder weights remained fixed throughout all experiments.

Before registration, the encoder features were reduced to $q=24$ channels using principal component analysis (PCA). The PCA basis was estimated from the fixed-image features and subsequently applied to both images:

\begin{equation}
	W 	= \operatorname{PCA}_q\!\left(F_{\mathrm{fix}}\right) \in \mathbb{R}^{C_{\mathrm{in}}\times q}, \qquad F^{q}_{\mathrm{fix}} = W^{\!\top}F_{\mathrm{fix}}, \qquad F^{q}_{\mathrm{mov}} 	= W^{\!\top}F_{\mathrm{mov}}.
	\label{eq:pca}
\end{equation}

The randomized singular value decomposition used for PCA was seeded for reproducibility.

Both representations were registered using \ca~\cite{convexadam2021}, which combines discrete correspondence search with instance-specific Adam refinement. For representation $\Phi$, the discrete stage evaluates

\begin{equation}
	\mathcal{C}_{\Phi}(x,u) = \parallel \Phi_{\mathrm{fix}}(x) - \Phi_{\mathrm{mov}}(x+u) \parallel_2^2
	\label{eq:cost}
\end{equation}
over integer displacements within a $\pm4$ voxel search region on a grid with spacing 6 voxels. Correspondences were regularized by coupled-convex relaxation with inverse consistency over 15 iterations.

The resulting displacement was refined using Adam optimization with a diffusion penalty $\lambda\,\mathbb{E}_x\|\nabla u(x)\|^2$. For the \mindssc stage, we used $\lambda=1.25$, 80 iterations, a learning rate of 1.0, parameter-grid spacing of 2, and three box filters of width 3. For the encoder stage, we used 500 iterations, a learning rate of 6.0, parameter-grid spacing of 1, and two box filters of width 7. The encoder-stage regularization weight was $\lambda=1$ for the monomodal datasets and $\lambda=2$ for \mrct. Because the two stages use different parameter-grid spacings, their regularization weights are stage-specific and should not be interpreted as directly comparable quantities. All solver settings were fixed within each dataset and were not retuned for individual coupling strategies.

The \mindssc stage used the reference \ca implementation, whereas the encoder stage used its parameterized interface to accept precomputed features. These interfaces are not numerically identical even when supplied with the same features. This implementation difference therefore remains a limitation of direct comparisons between MIND-only and encoder-based configurations, while the interface itself was held fixed across the corresponding ablation experiments. No solver parameter was optimized using the evaluated registration data, and no task-specific anatomical label was used during registration.

The reference configuration, \fm, combines the encoder and \mindssc representations in a sequential two-stage cascade shown in \figurename~\ref{fig:pipeline}(a-c). Let $\mathcal{S}[\Phi;v_0]$ denote one registration call using representation $\Phi$ initialized by displacement $v_0$. The cascade is defined as

\begin{equation}
	u_1 = \mathcal{S}\!\left[F^{q};0\right], \qquad u_2 = \mathcal{S}\!\left[\tilde{M}^{12};0\right], \qquad \varphi = 	\varphi_{\mathrm{affine}} \circ \varphi_{u_1} \circ \varphi_{u_2}, 
    \label{eq:cascade}
\end{equation}
where $u_1$ is estimated from the encoder features and $\tilde{M}^{12}$ denotes \mindssc computed after warping the moving image using $u_1$. The second stage therefore estimates a residual deformation from the coarsely aligned image pair.

For \dirlab and \acdc, centered initialization followed by rigid and affine registration was performed with SimpleITK before deformable registration. Mutual information was computed using 32 histogram bins and a seeded random sample of \pct{10} of voxels, with the anatomical masks applied to the metric. A centered initialization based on the fixed and moving mask regions was followed by rigid registration using shrink factors of 4 and 2 and an affine stage using shrink factors of 4, 2, and 1. The corresponding smoothing sigmas were 4 and 2 for the rigid stage and 4, 2, and 1 for the affine stage. Both stages used gradient descent for 300 iterations with optimizer scales estimated from physical shift. The resulting affine transformation was computed once for each image pair, cached, and shared across all configurations. For \abd, no additional affine registration was performed and $\varphi_{\mathrm{affine}}$ was the identity transformation.

The complete reference pipeline is illustrated in \figurename~\ref{fig:pipeline}. Encoder features were first used to estimate $u_1$. The moving image was then warped, \mindssc descriptors were recomputed from the fixed and warped moving images, and the residual deformation $u_2$ was estimated. The affine and two deformable transformations were subsequently composed to obtain the final transformation.

\begin{figure}[!tb]
	\centering
	\includegraphics[width=\linewidth]{flowchart.jpg}
	\caption{
		Overview of the reference \fm registration framework. For \dirlab and \acdc, a shared rigid-plus-affine pre-alignment $\varphi_{\mathrm{affine}}$ is applied to all configurations, whereas for \abd, $\varphi_{\mathrm{affine}}$ is the identity transformation. (a) Frozen encoder features are extracted slice-wise, reduced to $q=24$ channels using PCA, and used by \ca to estimate the initial deformation $u_1$. (b) The moving image is warped by $u_1$, after which \mindssc features are computed and a second \ca stage estimates the residual deformation $u_2$. (c) The affine and deformable transformations are composed to obtain the final registration.
	}
	\label{fig:pipeline}
\end{figure}

\subsection{Feature-coupling strategies}
\label{sec:combos}

Five alternative strategies were evaluated to determine whether modifying the interaction between the encoder and \mindssc representations improved registration relative to the reference sequential cascade. Each strategy modified one component of the reference pipeline while holding preprocessing, solver parameters, working resolution, affine initialization, and evaluation procedures fixed. The strategies are summarized in \tablename~\ref{tab:strategies}.

\begin{table}[t]
	\centering
	\small
	\caption{Feature-coupling strategies evaluated as modifications of the reference \fm cascade. $W$ denotes the PCA projection in Eq.~\ref{eq:pca}, $R$ is a seeded random projection of the same dimensions, $\mathcal{C}_{F^{q}}$ and $\mathcal{C}_{M^{12}}$ denote the encoder and \mindssc cost volumes, and $\gamma(x)\in[0,1]$ denotes a voxel-wise confidence weight. Components not explicitly modified were held fixed relative to the reference configuration.}
	\label{tab:strategies}
	
	\begin{tabular}{@{}lp{0.62\linewidth}@{}}
		\toprule
		Strategy & Modification \\
		\midrule
		
		Random projection &
		Replace the PCA projection $W$ with a seeded random projection $R$. \\
		
		Confidence gate &
		Weight the first-stage displacement as
		$u_1 \rightarrow \gamma \odot u_1$, where $\gamma$
		is derived from the sharpness of $\mathcal{C}_{F^{q}}$. \\
		
		Cost-volume fusion &
		Combine normalized encoder and \mindssc costs during the
		first stage as
		$\mathcal{C}_{F^{q}}/\overline{\mathcal{C}}_{F^{q}}
		+
		\mathcal{C}_{M^{12}}/\overline{\mathcal{C}}_{M^{12}}$. \\
		
		Composed fields &
		Estimate an additional encoder-derived displacement $u_{1b}$
		and compose it with $u_1$ before \mindssc refinement. \\
		
		MIND on the token grid &
		Apply a local self-similarity representation to $F^{q}$ on
		the encoder token grid before correspondence estimation. \\
		
		\bottomrule
	\end{tabular}
\end{table}

The random-projection configuration tested whether the data-adaptive PCA basis contributed to registration performance. Confidence gating attenuated the initial displacement in locations where the encoder correspondence cost provided weaker evidence for a unique match. Cost-volume fusion combined normalized encoder and \mindssc costs during the initial discrete search so that both representations contributed directly to correspondence selection. The composed-fields configuration introduced a second encoder-based registration after warping by $u_1$ and composed the resulting displacement before \mindssc refinement. MIND on the token grid applied local self-similarity to the reduced encoder representation before correspondence estimation.

The dimensionality-reduction procedure was examined separately from these coupling strategies. The reference fixed-image PCA was compared with a basis estimated jointly from the fixed and moving encoder features. On \abd, we additionally evaluated per-channel whitening and global scaling of the reduced encoder features to assess the sensitivity of the correspondence representation to feature normalization.

\section{Results}
\label{sec:results}

\subsection{Qualitative results}
\label{sec:results_qualitative}

\begin{figure}[!tbh]
	\centering
	\includegraphics[width=\linewidth]{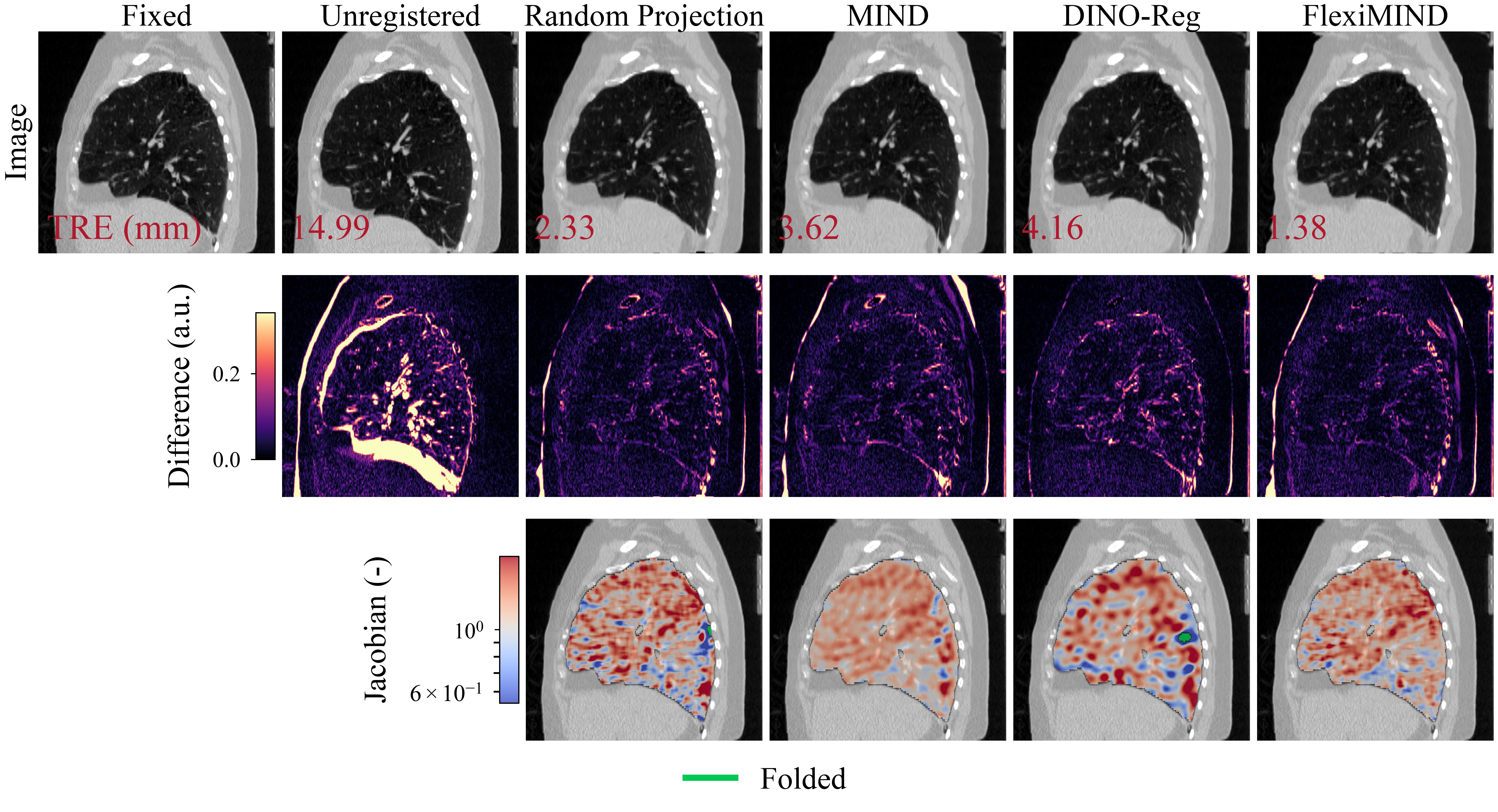}
	\caption{
		Qualitative registration result for \dirlab in the sagittal plane. Columns show the fixed image, unregistered moving image, and moving image	warped using random projection, \ca with \mindssc, \dinoreg, and \fm. Rows show the registered image, absolute intensity difference relative to	the fixed image, and Jacobian determinant of the estimated transformation. Voxels with non-positive Jacobian determinant are indicated in green.	Inset values report the case-specific \tre in millimeters.
		}
	\label{fig:qual_dirlab}
\end{figure}

\begin{figure}[!tbh]
	\centering
	\includegraphics[width=\linewidth]{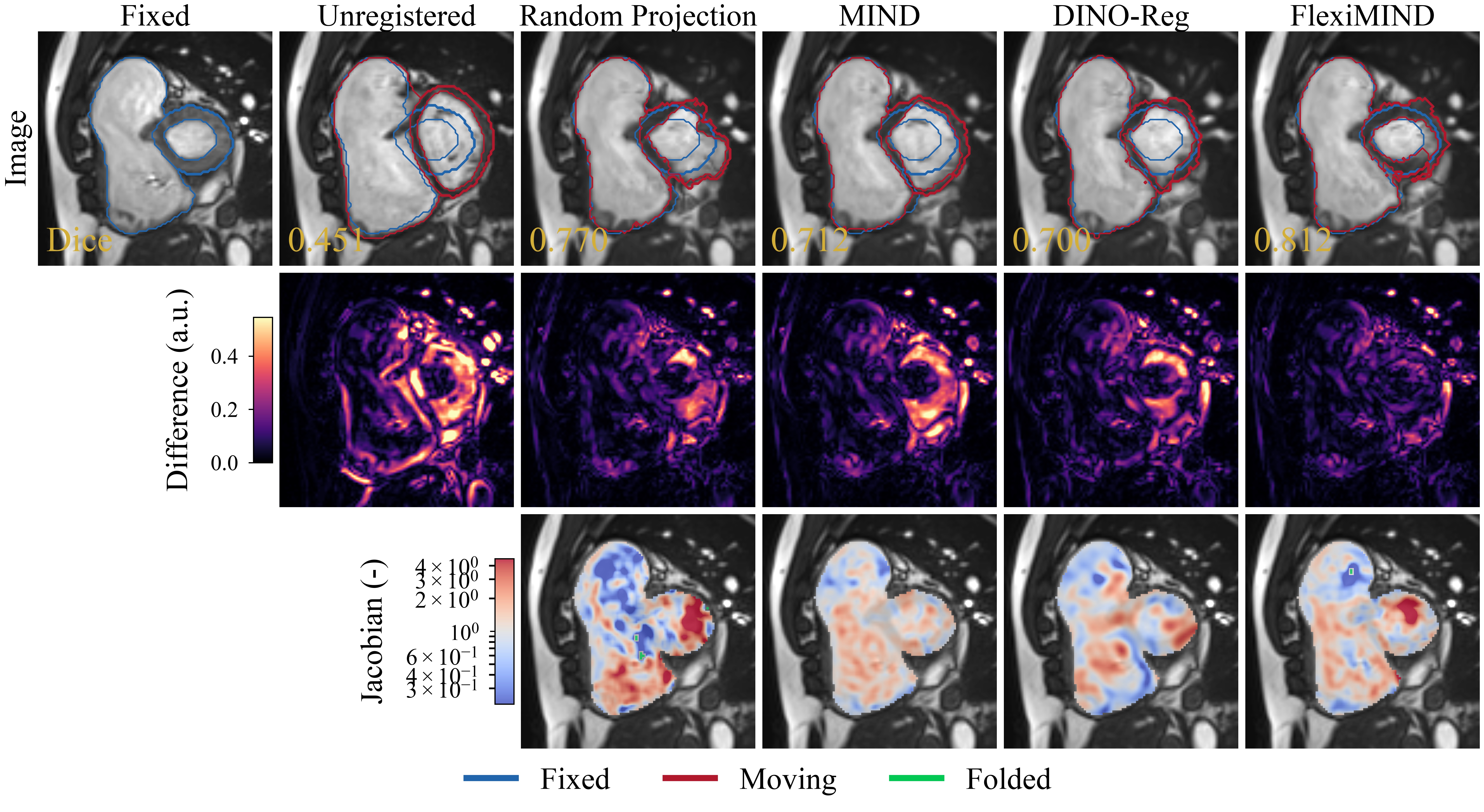}
	\caption{
		Qualitative registration result for \acdc in the short-axis plane. Columns follow Fig.~\ref{fig:qual_dirlab}. Fixed anatomical contours	are shown in blue and propagated moving contours in red. Rows show the registered image, absolute intensity difference relative to the fixed image, and Jacobian determinant of the estimated transformation. Inset values report the subject-averaged \dice for the displayed patient.
		}
	\label{fig:qual_acdc}
\end{figure}

\begin{figure}[!tbh]
	\centering
	\includegraphics[width=\linewidth]{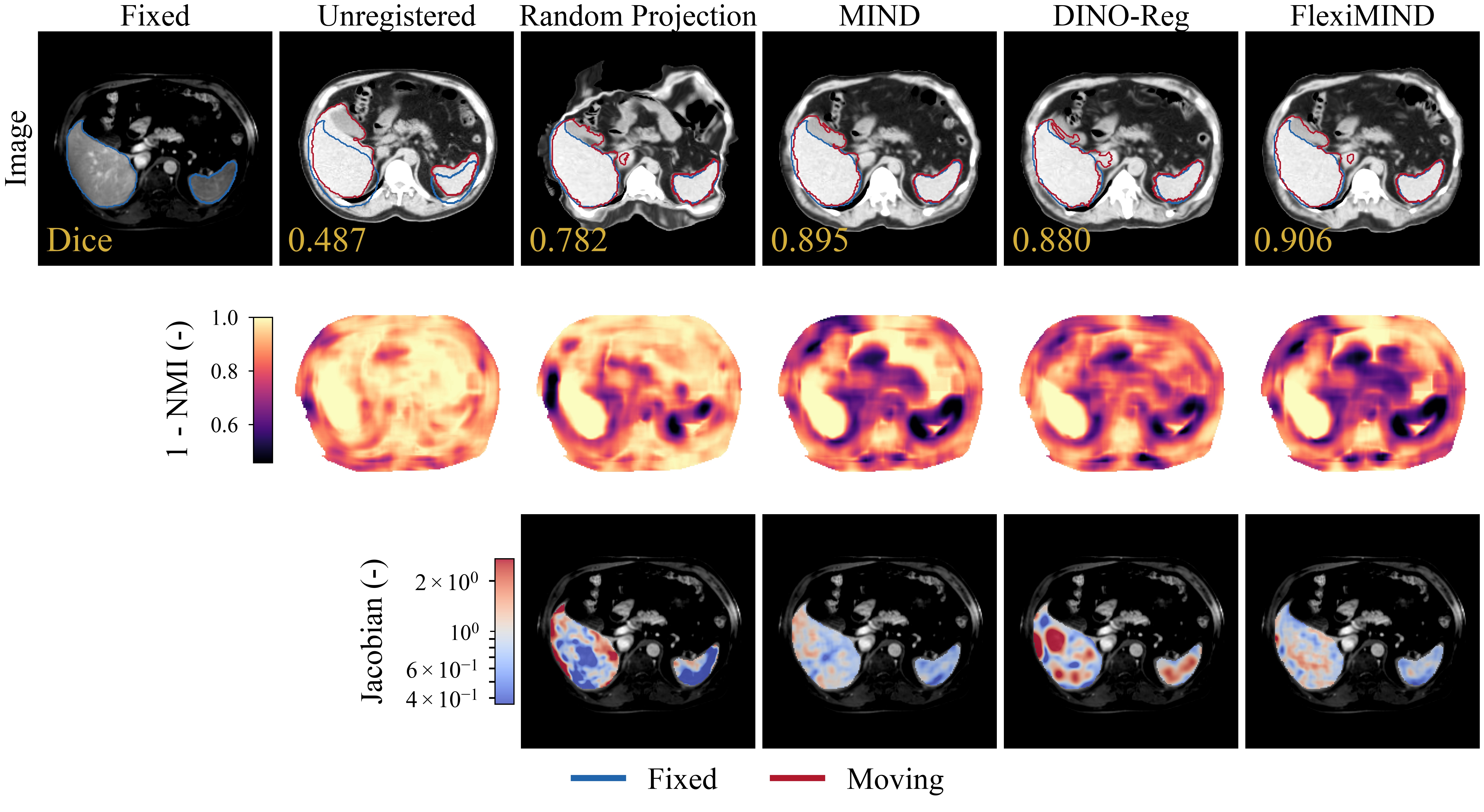}
	\caption{
		Qualitative registration result for Learn2Reg \abd pair 0008 in the axial plane. Columns follow Fig.~\ref{fig:qual_dirlab}. Fixed organ contours are shown in blue and propagated moving contours in red. The middle row shows $1-\mathrm{NMI}$ to account for the multimodal image pair, and the bottom row shows the Jacobian determinant of the estimated transformation. Inset values report subject-averaged \dice for the displayed pair.}
	\label{fig:qual_abdomenmrct}
\end{figure}

Figures~\ref{fig:qual_dirlab}-\ref{fig:qual_abdomenmrct} show representative registration results for the three modality pairings. For each dataset, the fixed image, unregistered moving image, and moving images warped by the evaluated configurations are shown together with a residual map and the Jacobian determinant of the estimated transformation. The random-projection strategy is included because it provides a representative example of how a change in the feature representation can affect both registration accuracy and deformation regularity.

Figure~\ref{fig:qual_dirlab} shows \dirlab case 8 in the sagittal plane, where the dominant superior-to-inferior respiratory motion is visible. The case-specific \tre was \mm{14.99} before registration, \mm{2.33} with random projection, \mm{3.62} with \ca and \mindssc, \mm{4.16} with \dinoreg, and \mm{1.38} with \fm. The corresponding Jacobian maps show different deformation behavior across the registered configurations, including a larger number of non-positive Jacobian determinants for \dinoreg than for \fm or \ca with \mindssc.

Figure~\ref{fig:qual_acdc} shows \acdc patient 140 in the short-axis plane. Subject-averaged \dice increased from 0.451 before registration to 0.770 with random projection, 0.712 with \ca and \mindssc, 0.700 with \dinoreg, and 0.812 with \fm. The propagated contours obtained with \fm show improved agreement with the fixed cardiac structures, while the Jacobian maps illustrate the spatial variation in the estimated deformation fields.

Figure~\ref{fig:qual_abdomenmrct} shows Learn2Reg \abd. Because the fixed and moving images are from different modalities, direct intensity differences are not informative and the residual row instead shows $1-\mathrm{NMI}$. Subject-averaged \dice increased from 0.487 before registration to 0.782 with random projection, 0.895 with \ca and \mindssc, 0.880 with \dinoreg, and 0.906 with \fm. The propagated organ contours show improved anatomical agreement after registration. Pair 0008 was a favorable case within the cohort and was selected because all four labeled organs were visible in a common axial plane.

\subsection{Quantitative results}
\label{sec:results_quantitative}

\begin{figure}[t]
	\centering
	\includegraphics[width=\linewidth]{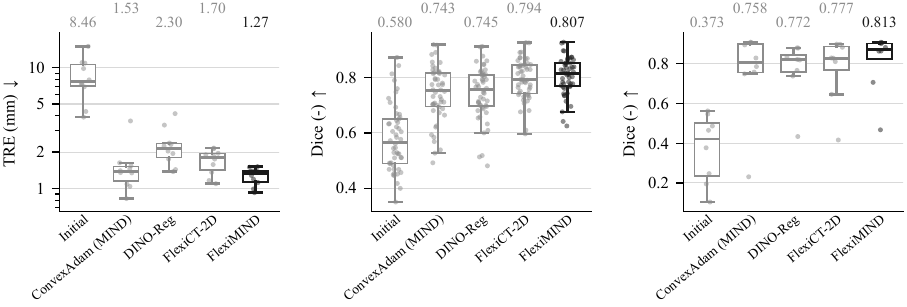}
	\caption{Per-case distributions of the registration metric for the configurations evaluated under the common study protocol. Individual cases are overlaid on the corresponding box plots to show the variability within each dataset.}
	\label{fig:distributions}
\end{figure}

\tablename~\ref{tab:main} summarizes registration accuracy, deformation-field regularity, and runtime for the configurations evaluated under the study protocol, together with selected published benchmark results. \figurename~\ref{fig:distributions} shows the corresponding per-case distributions.

On \dirlab, \fm achieved a mean \tre of \mm{1.2677}. Relative to the unregistered landmarks, the paired reduction in \tre was \mm{7.1935}, with a \pct{95} bootstrap interval of \mm{5.5467} to \mm{9.4182}. Compared with \dinoreg, the encoder-only configuration, and \ca with \mindssc, the reductions were \mm{1.0315}, \mm{0.4276}, and \mm{0.2623}, respectively. The corresponding paired \pct{95} intervals were \mm{0.6845} to \mm{1.6980}, \mm{0.2823} to \mm{0.5885}, and \mm{0.0004} to \mm{1.1455}. The interval for the comparison with \ca and \mindssc approached zero, indicating greater uncertainty in this difference than in the comparisons with the other configurations.

On \acdc, \fm achieved a subject-averaged \dice of 0.8067. The paired improvement was 0.2267 relative to the unregistered images, with a \pct{95} interval of 0.2049 to 0.2459. Compared with \ca and \mindssc, \dinoreg, and the encoder-only configuration, the improvements were 0.0640, 0.0621, and 0.0128, with corresponding intervals of 0.0521 to 0.0799, 0.0521 to 0.0744, and 0.0097 to 0.0164.

On \abd, \fm achieved a subject-averaged \dice of 0.8129 and an organ-averaged \dice of 0.8250. The paired improvement in subject-averaged \dice was 0.4397 relative to the unregistered images, 0.0546 relative to \ca with \mindssc, 0.0409 relative to \dinoreg, and 0.0361 relative to the encoder-only configuration. The corresponding \pct{95} intervals were 0.3452 to 0.5477, 0.0066 to 0.1345, 0.0044 to 0.0801, and 0.0214 to 0.0489. The wider intervals on this dataset reflect the smaller cohort of eight image pairs.

The \mindssc refinement added little computational cost to the encoder-based registration. Relative to the encoder-only configuration, the runtime of \fm increased by approximately 3.5\% on \dirlab, 2.9\% on \acdc, and 1.5\% on \abd. The complete cascade was approximately 3.8, 3.1, and 2.2 times faster than the reimplemented \dinoreg configuration on the same datasets, respectively.

\begin{table}[!tb]
	\centering
	\footnotesize
	\caption{
		Registration accuracy, deformation-field regularity, and runtime across the three modality pairings. Configurations evaluated in this study are reported as mean $\pm$ standard deviation with \pct{95} bootstrap confidence intervals. Published reference values report registration accuracy only and are shown separately within each dataset block. 		For \abd, subject-averaged \dice is reported with organ-averaged \dice in parentheses. Arrows indicate the direction of better registration accuracy. Runtime was measured on an NVIDIA RTX 6000 Ada GPU.
	}
	\label{tab:main}
	
	\resizebox{\textwidth}{!}{%
		\begin{tabular}{@{}llllll@{}}
			\toprule 	Dataset / metric & Configuration	& Accuracy 	& \pct{95} CI 	& \sdlogj & Runtime (s) \\
			\midrule
			
			\multirow{9}{*}{
				\shortstack[l]{
					\dirlab\\
					\ctct\\
					\tre (mm) $\downarrow$
				}
			}
			& Unregistered 	& $8.4612 \pm 3.1596$ 	& 6.6858 to 10.7230 	& --  & -- \\			
			& \ca
			& $1.5300 \pm 0.7348$ & 1.2402 to 2.3064 & 0.0430 & $29.2 \pm 10.8$ \\
			
			& \dinoreg & $2.2992 \pm 0.8165$ & 1.8995 to 2.9754 & 0.0633 & $608.8 \pm 148.8$ \\
			
			& Encoder only & $1.6954 \pm 0.3628$ & 1.4508 to 1.8954 & 0.0634 & $153.6 \pm 45.3$ \\
			
			& \fm & $\mathbf{1.2677} \pm 0.1950$ & 1.1305 to 1.3769 & 0.0643 & $159.0 \pm 47.1$ \\
			
			\cdashline{2-6}
			\addlinespace[0.5em]
			& \multicolumn{5}{l}{\textit{Published reference values}} \\
			
			& AD-RegNet~\cite{adregnet} & 1.51 & & & \\
			
			
			& Observer uncertainty~\cite{castillo2009dirlab} & 0.88 & & & \\
			
			\addlinespace[0.7em]
			\hline
			\addlinespace[0.7em]
			
			\multirow{8}{*}{
				\shortstack[l]{
					\acdc\\
					\mrmr\\
					\dice $\uparrow$
				}
			}
			& Unregistered & $0.5801 \pm 0.1189$ & 0.5486 to 0.6142 & -- & -- \\
			
			& \ca
			& $0.7428 \pm 0.0996$ & 0.7125 to 0.7681 & 0.0582 & $19.7 \pm 2.9$ \\
			
			& \dinoreg & $0.7447 \pm 0.0950$ & 0.7151 to 0.7681 & 0.0966 & $324.9 \pm 31.4$ \\
			
			& Encoder only & $0.7939 \pm 0.0718$ & 0.7726 to 0.8121 & 0.0877 & $103.4 \pm 8.7$ \\
			
			& \fm
			& $\mathbf{0.8067} \pm 0.0655$
			& 0.7870 to 0.8231
			& 0.0960
			& $106.4 \pm 9.0$ \\
			
			\cdashline{2-6}
			\addlinespace[0.5em]
			
			& \multicolumn{5}{l}{\textit{Published reference values}} \\
			
			& Slicer Network~\cite{slicernet} (label-free)
			& 0.7952
			& & & \\
			
			& MemWarp~\cite{memwarp} (uses labels)
			& 0.8961
			& & & \\
			\addlinespace[0.7em]
			\hline
			\addlinespace[0.7em]
			
			\multirow{8}{*}{
				\shortstack[l]{
					Learn2Reg \\\abd\\
					\mrct\\
					\dice $\uparrow$
				}
			}
			& Unregistered
			& $0.3732 \pm 0.1612$ (0.3709)
			& 0.2500 to 0.4757
			& --
			& -- \\
			
			& \ca
			& $0.7583 \pm 0.2085$ (0.7777)
			& 0.5152 to 0.8514
			& 0.1069
			& $4.6 \pm 0.6$ \\
			
			& \dinoreg
			& $0.7720 \pm 0.1358$ (0.7850)
			& 0.6143 to 0.8349
			& 0.1972
			& $413.3 \pm 54.0$ \\
			
			& Encoder only
			& $0.7767 \pm 0.1560$ (0.7880)
			& 0.6102 to 0.8546
			& 0.1272
			& $184.7 \pm 6.8$ \\
			
			& \fm
			& $\mathbf{0.8129} \pm 0.1442$ (\textbf{0.8250})
			& 0.6513 to 0.8829
			& 0.1448
			& $187.4 \pm 7.1$ \\
			
			\cdashline{2-6}
			\addlinespace[0.5em]
			& \multicolumn{5}{l}{\textit{Published reference values}} \\
			
			& \ca
			& (0.722)
			& & & \\
			
			& \dinoreg~\cite{dinoreg2025}
			& (0.785)
			& & & \\
			
			\bottomrule
		\end{tabular}
	}
\end{table}

Figure~\ref{fig:distributions} shows that the improvement was not uniform across cases. On \dirlab, much of the difference between \fm and \ca with \mindssc arose from a subset of cases, whereas the distributions on \acdc showed a more consistent shift toward higher \dice. The \abd results showed substantially greater inter-case variability, consistent with the wider bootstrap intervals observed for this dataset.

Among the published results considered here, \fm achieved lower \tre on \dirlab than the learned methods included in the comparison, including AD-RegNet at \mm{1.51}~\cite{adregnet}.  On \acdc, the subject-averaged \dice of 0.8067 exceeded the published label-free Slicer Network result of 0.7952~\cite{slicernet}. Methods trained using anatomical labels achieved higher values, including MemWarp at 0.8961~\cite{memwarp}, and were treated separately because they used task-specific supervision. On \abd, the organ-averaged \dice of 0.8250 exceeded the 0.785 reported for \dinoreg~\cite{dinoreg2025}. Our reimplementation of \dinoreg produced an organ-averaged \dice of 0.7850.

\begin{figure}[t]
	\centering
	\includegraphics[width=0.9\linewidth]{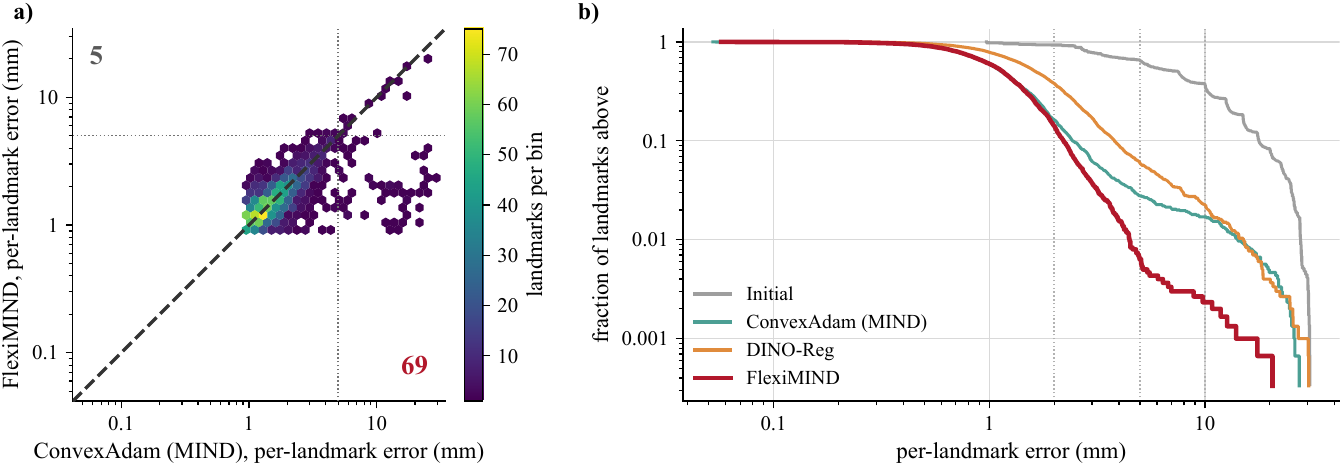}
	\caption{
		Landmark-level comparison on \dirlab using all 3000 annotated landmarks. (a) Per-landmark error of \fm versus \ca with \mindssc; the dashed line indicates equal error and dotted lines mark \mm{5}. (b) Fraction of landmarks with error exceeding the indicated threshold for the evaluated configurations.	Both panels use unrounded landmark errors and are therefore not directly comparable with the voxel-rounded \tre values in Table~\ref{tab:main}.}
	\label{fig:landmarks}
\end{figure}

The \dirlab result was further examined at the individual-landmark level to determine whether the reduction in mean \tre reflected a uniform improvement or a change in the error distribution. \figurename~\ref{fig:landmarks} compares unrounded errors for all 3000 landmarks. \fm produced a lower error than \ca with \mindssc for 1496 landmarks (\pct{49.9}), and the pooled median errors were similar at \mm{1.1578} for \fm and \mm{1.1437} for \ca with \mindssc. The main difference occurred in the upper tail of the distribution. The proportion of landmarks with errors above \mm{5} decreased from \pct{2.73} with \ca and \mindssc to \pct{0.60} with \fm, while errors above \mm{10} decreased from \pct{1.67} to \pct{0.20}. In paired terms, 69 landmarks moved from above to below the \mm{5} threshold, whereas five changed in the opposite direction.

Deformation-field regularity was evaluated to determine whether differences in registration accuracy were accompanied by changes in deformation behavior. No configuration produced a folding under the official Learn2Reg threshold of $\det J_u \leq -3$. The proportion of voxels with non-positive Jacobian determinant was more informative. For \fm, these proportions were \pct{0.0055} on \dirlab, \pct{0.25} on \acdc, and \pct{0.14} on \abd. In contrast, the random-projection strategy produced \pct{0.059}, \pct{0.31}, and \pct{28.0}, respectively, showing that similar registration objectives can produce substantially different local deformation behavior.

The \sdlogj of \fm was 0.0643 on \dirlab, 0.0960 on \acdc, and 0.1448 on \abd. \ca with \mindssc produced smoother fields on all three datasets, with corresponding values of 0.0430, 0.0582, and 0.1069. Relative to \dinoreg, \fm had a slightly higher \sdlogj on \dirlab (0.0643 versus 0.0633), a similar value on \acdc (0.0960 versus 0.0966), and a lower value on \abd (0.1448 versus 0.1972). These results indicate that the observed accuracy differences were accompanied by dataset-dependent changes in deformation regularity rather than by a uniform increase or decrease in field smoothness.

Because the additive offset in Eq.~\ref{eq:sdlogj} compresses the reported statistic, an offset-free sensitivity analysis was also performed. The direction of the regularity change was preserved in 17 of the 18 strategy-by-dataset comparisons.

\section{Ablation study}
\label{sec:ablation}

\begin{figure}[!tbh]
	\centering
	\includegraphics[width=0.8\linewidth]{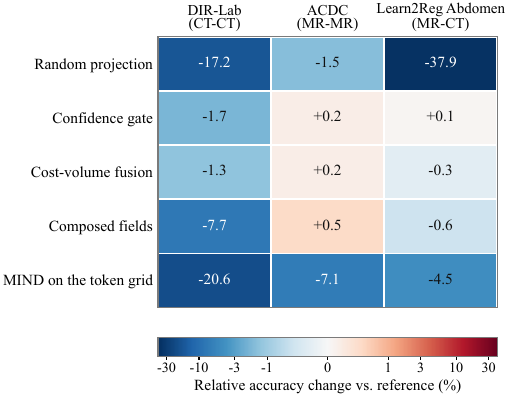}
	\caption{Relative change in registration accuracy for the five feature-coupling strategies compared with the reference \fm cascade. The sign is normalized across metrics so that positive values indicate improved registration and negative values indicate reduced registration accuracy.}
	\label{fig:ablationgrid}
\end{figure}

The five feature-coupling strategies defined in \tablename~\ref{tab:strategies} were evaluated against the reference \fm cascade on all three datasets. \tablename~\ref{tab:ablation} reports registration accuracy and deformation-field regularity, while \figurename~\ref{fig:ablationgrid} summarizes the relative change in registration accuracy with the sign normalized so that positive values indicate improvement for all three metrics.

None of the five strategies improved on \fm for \ctct registration. On \dirlab, cost-volume fusion, confidence gating, composed fields, random projection, and MIND on the token grid increased \tre by \mm{0.0161}, \mm{0.0221}, \mm{0.0978}, \mm{0.2177}, and \mm{0.2617}, respectively. These changes correspond to relative degradations of approximately \pct{1.3}, \pct{1.7}, \pct{7.7}, \pct{17.2}, and \pct{20.6} compared with the reference cascade.

Across all 15 strategy-by-dataset comparisons, only four produced higher registration accuracy than \fm, and all improvements were small. Confidence gating increased \dice by 0.0015 on \acdc and 0.0004 on \abd, cost-volume fusion increased \dice by 0.0015 on \acdc, and composed fields increased \dice by 0.0043 on \acdc. No strategy improved accuracy consistently across the three registration settings.

The degradations were substantially larger than the observed improvements. Random projection increased \tre by \pct{17.2} on \dirlab and reduced subject-averaged \dice on \abd from 0.8129 to 0.5047, a relative decrease of approximately \pct{37.9}. MIND on the token grid increased \tre by \pct{20.6} on \dirlab and reduced \dice by 0.0569 on \acdc and 0.0370 on \abd. Composed fields produced the largest positive change, increasing \dice on \acdc from 0.8067 to 0.8110, but reduced accuracy on both \dirlab and \abd. Overall, tighter coupling of the two representations did not provide a consistent improvement over the sequential cascade.

\begin{table}[!tbh]
	\centering
	\footnotesize	
	\caption{Ablation results for the five feature-coupling strategies and the reference \fm cascade. Registration accuracy is reported as \tre in millimeters for \dirlab and subject-averaged \dice for \acdc and \abd. \sdlogj characterizes deformation-field regularity, with lower values indicating smoother fields.}
	\label{tab:ablation}
	\begin{tabular}{@{}lcccccc@{}}
		\toprule
		& \multicolumn{2}{c}{\dirlab} & \multicolumn{2}{c}{\acdc} & \multicolumn{2}{c}{\abd} \\
		\cmidrule(lr){2-3}
		\cmidrule(lr){4-5}
		\cmidrule(lr){6-7}
		Configuration & \tre (mm) & \sdlogj & \dsc & \sdlogj & \dsc & \sdlogj \\
		\midrule
		\fm (reference)        & \textbf{1.2677} & 0.0643 & 0.8067 & 0.0960 & \textbf{0.8129} & 0.1448 \\
		Random projection      & 1.4854 & 0.0824 & 0.7948 & 0.1267 & 0.5047 & 0.4022 \\
		Confidence gate        & 1.2898 & 0.0653 & 0.8082 & 0.0987 & \textbf{0.8133} & 0.1451 \\
		Cost-volume fusion     & 1.2838 & 0.0651 & 0.8082 & 0.0973 & 0.8104 & 0.1451 \\
		Composed fields        & 1.3655 & 0.0825 & \textbf{0.8110} & 0.1187 & 0.8082 & 0.1863 \\
		MIND on the token grid & 1.5294 & 0.0446 & 0.7498 & 0.0609 & 0.7759 & 0.1088 \\
		\bottomrule
	\end{tabular}
\end{table}

The dimensionality-reduction procedure was also examined independently. The reference configuration estimated the PCA basis from the fixed-image features and applied the same projection to both images. Replacing this procedure with a PCA basis fitted jointly to the fixed and moving features had little effect on the two monomodal datasets. On \dirlab, joint PCA produced a mean \tre of \mm{1.2646}, compared with \mm{1.2677} for fixed-image PCA, while on \acdc the corresponding \dice values were 0.8083 and 0.8067. In contrast, joint PCA substantially reduced performance on the multimodal \abd dataset, where organ-averaged \dice decreased from 0.8250 to 0.6524 and \sdlogj increased from 0.1448 to 0.3444. This result supports the use of a projection estimated from a single image when the two inputs arise from different modalities.

Additional normalization of the reduced encoder features was also evaluated on \abd. Per-channel whitening produced an organ-averaged \dice of 0.5942 with \sdlogj of 1.3646, while global scaling produced a \dice of 0.7039 with \sdlogj of 1.2515. Both configurations therefore produced substantially less regular deformation fields together with lower registration accuracy than the reference configuration.

\begin{figure}[!tbh]
	\centering
	\includegraphics[width=\linewidth]{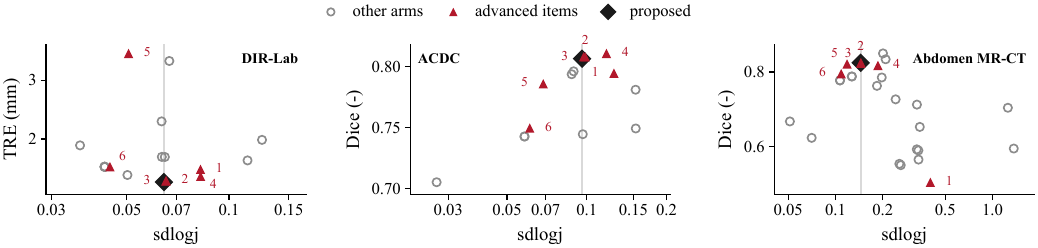}
	\caption{
		Relationship between deformation-field regularity and registration accuracy across the evaluated configurations. Open circles denote other measured	configurations, red triangles denote the five feature-coupling strategies, and the black diamond denotes \fm. Coupling strategies are numbered as: (1) random projection, (2) confidence gate, (3) cost-volume fusion,	(4) composed fields, and (5) MIND on the token grid. The vertical line	indicates the \sdlogj of \fm. Larger deviations from the regularity of the reference configuration were generally associated with larger changes in registration accuracy.}
	\label{fig:regularity}
\end{figure}

Figure~\ref{fig:regularity} relates registration accuracy to \sdlogj across the evaluated configurations. The two strategies that produced deformation regularity closest to the reference cascade, confidence gating and cost-volume fusion, also produced the smallest changes in registration accuracy. Confidence gating changed \sdlogj by less than approximately \pct{3} across all three datasets, while its accuracy remained close to that of \fm. Cost-volume fusion showed a similar pattern.

Larger changes in deformation regularity were generally accompanied by larger reductions in registration accuracy. Random projection increased \sdlogj from 0.0643 to 0.0824 on \dirlab and from 0.1448 to 0.4022 on \abd, together with substantial losses in registration accuracy. Composed fields also increased \sdlogj on all three datasets and improved accuracy only on \acdc. In the opposite direction, MIND on the token grid produced smoother fields than \fm on all three datasets, with \sdlogj decreasing from 0.0643 to 0.0446 on \dirlab, from 0.0960 to 0.0609 on \acdc, and from 0.1448 to 0.1088 on \abd, while registration accuracy decreased in every setting.

These results indicate that the observed effects of feature coupling were closely associated with changes in deformation-field regularity. Importantly, neither increasing nor decreasing regularity relative to the reference cascade consistently improved registration accuracy. The alternative coupling strategies therefore provided no evidence of a robust correspondence advantage over the sequential combination of encoder features followed by \mindssc refinement.

\section{Discussion}
\label{sec:discussion}

The reference \fm cascade achieved strong registration accuracy across the three evaluated modality pairings and outperformed the comparator configurations measured under the same protocol. In contrast, none of the five alternative feature-coupling strategies produced a consistent improvement over the sequential cascade. The few positive changes were small and dataset dependent, whereas several strategies produced substantially larger reductions in accuracy. These findings suggest that the different spatial properties of \mindssc and pretrained encoder features do not, by themselves, guarantee that tighter integration of the two representations will provide more informative correspondences. In the evaluated settings, their most effective use was instead sequential, with encoder features estimating the initial deformation and \mindssc refining the residual alignment.

The behavior of the coupling strategies can be interpreted in the context of the registration solver. The discrete search first identifies candidate correspondences from the feature cost surface, after which continuous optimization refines the resulting displacement under spatial regularization. A modification to the feature representation can therefore affect both the selected correspondence and the regularity of the estimated deformation. In our experiments, strategies that preserved deformation regularity close to that of the reference cascade, particularly confidence gating and cost-volume fusion, also produced only small changes in registration accuracy. Larger changes in regularity were generally associated with larger reductions in accuracy. Random projection and composed fields produced less regular deformation fields and degraded performance in most settings, whereas MIND on the token grid produced smoother fields but consistently reduced registration accuracy. These observations do not establish that regularity caused the changes in accuracy, but they show that the effect of feature coupling cannot be interpreted independently of the deformation field it produces.

This distinction is important when evaluating feature-fusion strategies for deformable registration. If a proposed representation changes registration accuracy while also substantially changing deformation regularity, the source of the improvement cannot be assigned uniquely to additional correspondence information. Comparisons are therefore more informative when the solver, working resolution, dimensionality reduction, and regularization are controlled, and when deformation regularity is reported together with registration accuracy. Our results do not imply that previously reported improvements from feature fusion are caused by changes in regularization. Rather, they demonstrate that changes in correspondence representation and changes in deformation behavior can occur simultaneously and should be separated when possible.

The \dirlab landmark analysis provides additional insight into the role of the encoder stage. \fm did not improve the majority of individual landmarks, and the median landmark error remained similar to that of \ca with \mindssc. The reduction in mean \tre was instead driven primarily by a reduction in the upper tail of the error distribution, with substantially fewer landmarks exceeding \mm{5} and \mm{10}. This suggests that the encoder representation contributes mainly by reducing large correspondence failures rather than by uniformly improving local precision. A local descriptor such as \mindssc may be insufficient to distinguish anatomically similar structures when the correct correspondence requires broader spatial context, whereas the larger contextual field of a pretrained encoder may help resolve these ambiguities. The subsequent \mindssc stage can then refine the alignment using local three-dimensional structural information. This interpretation is consistent with the observed benefit of the sequential ordering, although the present experiments do not directly isolate the anatomical structures for which each representation is most informative.

The dimensionality-reduction experiments further show that the representation used for correspondence estimation is influenced not only by the encoder but also by how its features are projected. Fitting the PCA basis jointly to the fixed and moving features had little effect on the two monomodal datasets but substantially reduced accuracy and increased deformation irregularity on \mrct. This suggests that a joint basis can become sensitive to modality-dependent differences when the two feature distributions differ substantially. Estimating the basis from the fixed image and applying the same transformation to both images avoids introducing such pair-dependent information into the projection. The poor results obtained with additional whitening and global scaling on \abd further indicate that seemingly simple changes in feature normalization can substantially alter both the correspondence cost and the resulting deformation field. Dimensionality reduction should therefore be treated as part of the registration representation rather than as a purely computational preprocessing step.

The computational results also support the sequential design. Adding the \mindssc refinement stage increased runtime by only approximately 1.5--3.5\% relative to the encoder-only configuration across the three datasets, while improving registration accuracy in each setting. Under the same hardware and experimental environment, the complete cascade was approximately 2.2--3.8 times faster than the reimplemented \dinoreg configuration. The additional local refinement therefore provided an accuracy improvement with relatively little computational overhead beyond the encoder-based registration. These runtime comparisons apply to the implementations and hardware used in this study and should not be interpreted as general speed comparisons with implementations evaluated under different computational environments.

Comparison with previous work provides additional context for these findings. Our reimplementation of \dinoreg produced an abdominal result close to its published value~\cite{dinoreg2025}, supporting the consistency of the implementation used for the comparison. DINO-Reg also reported that \ca with \mind could outperform its encoder-based approach in lung registration~\cite{dinoreg2025}, where local contour and structural information may remain particularly informative. Our \ctct results are consistent with the importance of retaining the hand-crafted descriptor rather than replacing it entirely with pretrained features. The use of pretrained encoder features therefore reduces large correspondence errors in our framework but does not eliminate the remaining gap to the strongest classical methods on this dataset.

The regularity analysis also highlights limitations of relying on a single deformation-field statistic. The official Learn2Reg folding criterion uses the shifted quantity $\det J_u+3$ and therefore does not identify all voxels with non-positive Jacobian determinant. In our experiments, the official folding count was zero even for configurations that contained non-positive Jacobians and visibly irregular deformation fields. Similarly, the additive offset used in the official \sdlogj formulation compresses differences between fields and can change the ordering of configurations with similar regularity. These metrics remain useful for comparison with results evaluated using the same Learn2Reg software, but they are less suitable for fine-grained interpretation of deformation behavior. Reporting the proportion of non-positive Jacobian determinants and examining an offset-free regularity measure provided complementary information in the present study.

Several limitations should be considered when interpreting these results. The \dirlab and \abd cohorts contain only ten and eight image pairs, respectively, resulting in relatively wide uncertainty intervals, and we therefore relied on paired effect estimates and bootstrap confidence intervals rather than formal hypothesis testing. Feature extraction was performed independently on axial slices using a two-dimensional encoder, which provides limited explicit through-plane information; this limitation is particularly relevant to \acdc, where the native through-plane spacing ranges from \SIrange{5}{10}{\milli\meter}. In addition, the \mindssc and encoder stages use different interfaces of the \ca implementation, which prevents a completely isolated comparison of the two feature representations. Finally, runtime was measured on a single hardware platform, and the reported computational differences may change with implementation and hardware.

These findings motivate several directions for further investigation. Three-dimensional or multi-plane pretrained representations could provide broader contextual information without sacrificing through-plane correspondence, while hand-crafted descriptors with larger three-dimensional receptive fields could test whether the long-range disambiguation provided by the encoder can be achieved without a separate semantic stage. Future evaluations of feature coupling would also benefit from explicitly regularity-matched comparisons, in which correspondence representations are compared at similar deformation regularity rather than under a single fixed smoothness parameter. Such experiments would provide a more direct test of whether combining representations adds correspondence information beyond the effect of changing the deformation operating point.

\section{Conclusion}
\label{sec:conclusion}

We developed a training-free and label-free deformable registration framework that combines pretrained encoder features with \mindssc in a sequential \ca-based cascade. Across \ctct, \mrmr, and \mrct registration, tighter feature coupling did not consistently improve accuracy over the sequential design, while changes in performance were closely associated with deformation-field regularity.These findings support a scale-separated registration strategy in which contextual encoder features resolve coarse anatomical correspondence and local self-similarity descriptors provide spatial refinement, while emphasizing the importance of deformation regularity when evaluating feature-coupling strategies.

\section*{Conflicts of interest}
There are no conflicts of interest declared by the authors.
\section*{Acknowledgment}
This research is supported in part by the National Institutes of Health under Award Numbers 1R01HL183223, R01DE033512 and R01CA272991.

\section*{Data availability}

All datasets used in this study are publicly available. The \dirlab 4-D CT dataset, the \acdc cardiac MRI dataset, and the Learn2Reg \abd MR-CT dataset can be accessed through their respective public repositories and challenge websites. No private or institutional patient data were used in this study.

\printbibliography
\end{document}